# Visualizing the Flow of Discourse with a Concept Ontology

Baoxu Shi
University of Notre Dame
bshi@nd.edu

Tim Weninger
University of Notre Dame
tweninge@nd.edu

## ABSTRACT
Understanding and visualizing human discourse has long being a challenging task. Although recent work on argument mining have shown success in classifying the role of various sentences, the task of recognizing concepts and understanding the ways in which they are discussed remains challenging. Given an email thread or a transcript of a group discussion, our task is to extract the relevant concepts and understand how they are referenced and re-referenced throughout the discussion. In the present work, we present a preliminary approach for extracting and visualizing group discourse by adapting Wikipedia's category hierarchy to be an external concept ontology. From a user study, we found that our method achieved better results than 4 strong alternative approaches, and we illustrate our visualization method based on the extracted discourse flows.

## 1 INTRODUCTION
Language has long been one of the most efficient forms of communication between people. Technology that can parse and extract information from these conversations currently exists and operates with reasonable accuracy; however, there is a gap in our ability to understand and visualize these conversational statements. Current and previous work in the analysis of news articles and social posts have demonstrated the ability to extract and quantify written ideas [1, 2, 5]; however, these tools operate over large text or news corpora, so they are not able to discover concept flows of individual conversations (or documents). Related work in application-oriented natural language processing aims to extract named entities or important concepts and entities from sentences. Although Named Entity Recognition may be able to extract high-quality entities, which could be viewed as concepts, they are usually limited to a few entity types.

The goal of the present work is different. Here we transform a group conversation into a network over concepts in order to visualize the concept flows so that we might better understand the latent communication patterns and group dynamics. Our key insight is to treat human group conversations as trails over a graph of concepts. With this perspective, an individual's ideas as expressed through language can be mapped to explicit entities or concepts, and, therefore, a single argument can be treated as a path over the graph of concepts.

We overcome the limitations mentioned above by distilling a high-quality concept ontology from Wikipedia and using its entity surface forms to detect concepts in human discourse. We then find concept flows by computing sentence similarities using a joint text and concept similarity. The code and data are available at https://github.com/bxshi/DiscourseVisualization.

## 2 DISCOURSE GRAPHIFICATION
In the present work, we assume each Wikipedia article represents a unique concept and further treat the categories that an article belongs to as more general concepts. This solution assumes that the Wikipedia category hierarchy is a clean ontology. This is not the case. So the first step is to perform some pre-processing to transform the Wikipedia category hierarchy into a useful ontology.

We use the October 2017 English dump of all Wikipedia articles and categories. We begin by removing all maintenance, tracking, chronological and list-like pages such as Articles to be split and 1880 deaths, etc. This results in a graph rooted at the category-page Main Topic Classifications with $976,163$ category nodes, $1,901,706$ fine-to-coarse concept edges, and $11,967,618$ unique leaf-concepts corresponding to Wikipedia articles. Each article belongs to $4.75$ categories on average (mean). A snippet of this concept ontology is illustrated in Fig. 1.

The next step is to extract concepts from transcripts of group conversations and link the sentences to form concept flows over the ontology. To extract concepts from the discourse text, we simply match the surface forms of concepts $\mathbf{E}_i$ from the $i^{\text{th}}$ sentence $\mathbf{S}_i$ within in transcript $\mathbf{D}$ against the Wikipedia article titles (leaves in the Wikipedia concept ontology). Because each concept-leaf is associated with one or more parent and ancestor concepts, we say

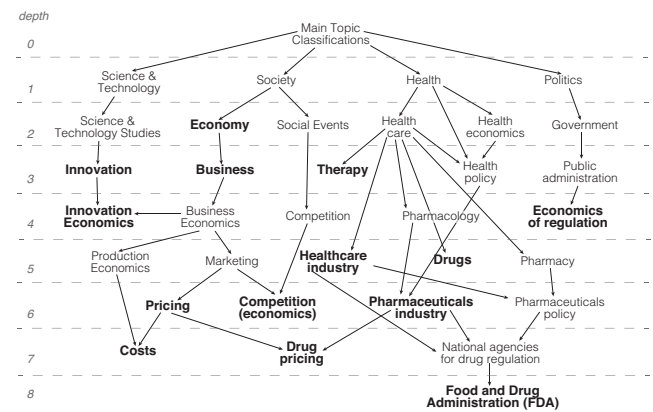

**Figure 1: Illustration of a snippet of the concept graph. Bold texts are extracted concepts recognized as important in the Intelligence[2] debate referenced in Fig. 2.**

| Seq | Speaker | Sentence |
|---|---|---|
| $A_1$ | J. Donvan | if you think of how many times you've taken **antibiotics**, casually saving yourself from **death** by **infection**. |
| $A_2$ | J. Donvan | if you think about **vaccines** and the way they have shielded you against deadly and disabling **disease**, if you think about the fact that **hiv infection** is no longer a **death** sentence, |
| $B_1$ | P. Howard | some **patients**, particularly those with serious **chronic illnesses**, are paying too much out of pocket for their **medicines**, and we need to find a **solution** for that |
| $C_1$ | E. Emanuel | but we don't want those **drug prices** of **$150,000** per year to take those **drugs** |
| $B_2$ | E. Emanuel | it took a **disease** that was a **chronic disease** – but blasted off and killed people in six months and basically made people live a very long time with the **disease**. |
| $C_2$ | E. Emanuel | there are multiple **drugs** out there on the market that are about **$150,000** per year, don't **cure** anyone, ameliorate the **disease**, but are hugely **expensive**. |
| $B_3$ | L. Reilly | today, it's **chronic disease** and **treating patients** with **chronic disease** that are responsible for 90 percent of all **healthcare costs**. |
| $C_3$ | E. Emanuel | the problem is we have these super high **drug prices**, 150,000, $300,000 drugs that don't **cure** anyone, and they're still exorbitantly **expensive**. |
| $C_4$ | P. Howard | micromanage who gets what **price** and who can do what, and that is one of the biggest things standing in front of us, especially the **fda**'s **drug regulations** that make it so difficult to **innovate** and so **expensive** to **innovate**. |
| $C_5$ | P. Howard | and if we change **pricing** without changing how we **innovate**, all we're going to wind up is with fewer **drugs** and a **drug** that you don't have for a serious **disease** is infinitely **expensive** because you can't buy it. |
| $C_6$ | E. Emanuel | we're here to debate **drug pricing** and **drugs** outrageously high **prices**. |

↑ Debate Snippet    Conversation Diagram↓

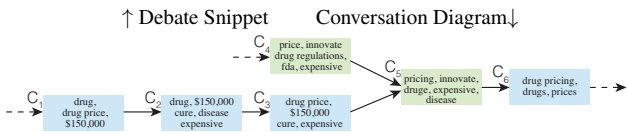

Figure 2: A partial example of the Intelligence Squared debate *Blame Big Pharma for Out-of-Control Health Care Costs*. Sentences are color-coded by speaker and have assigned labels. Bold texts are extracted concepts shown in Fig. 1. At bottom is a flow diagram over concepts mentioned during the debate.

that each sentence is associated with a concept tree as an induced subgraph $\mathbf{C}_i$ from the concept ontology.

Next we need to link concepts across sentences in the discourse. This requires some notion of concept similarity. There are many ways to do this. We initially tried to adapt the Jaccard coefficient, but this did not work well because it fails to consider the concept granularity and instead treats all concepts, regardless their position in the ontology, equally. To properly weight the concepts, we apply TF-IDF weighting to the extracted concepts by treating them as "words" and the sentences as "documents". We further define the concept feature vector $V_i$ of the $i^{th}$ sentence as

$$\mathbf{V}_i = \left\{ \mathbb{I}(c_k \in \mathbf{C}_i) \times \left(1 + \log \frac{N}{\sum_{j=1}^{N} \mathbb{I}(c_k \in \mathbf{C}_j)}\right) \Big| k \in \{1 \ldots m\} \right\}, \quad (1)$$

in which $m$ and $N$ are the total number of concepts and sentences respectively. We can get the word feature vector $\mathbf{U}_i$ using the same method. Putting these together, we now define the sentence similarity as the combination of the word and concept cosine similarities:

$$\text{sim}(\mathbf{S}_i, \mathbf{S}_j) = \theta(\mathbf{V}_i, \mathbf{V}_j) + \theta(\mathbf{U}_i, \mathbf{U}_j). \quad (2)$$

Using Eq. 2 we can now construct concept flows by linking similar sentences together and highlighting important words and concepts in the sentences. For each sentence $\mathbf{S}_i \in \mathbf{D}$, we find the most similar sentence $\mathbf{S}_j$ in which $i < j$ and illustrate the concept relationships using a concept network as shown in Fig. 2.

## 3 EXPERIMENTS

We performed a user study to evaluate how well this model captures sentence-level semantic similarities. We compared our model with the results of TopicFlow (LDA) [3], word overlap baseline, averaged sentence embeddings from GloVe [4], and a text-only version of our model using only $\theta(\mathbf{U}_i, \mathbf{U}_j)$ from Eq. 2. Our dataset consisted of four debates from intelligence[2] covering Politics, Health, Science and Economics. We randomly selected 20 sentences from each debate and used the methods mentioned above to find the most similar sentence for each selected sentence. Then for each sentence pair, we asked 10 human annotators to rank the similarity on a 0 to 4 Likert Scale. The results are in Fig. 3 with 95% confidence intervals.

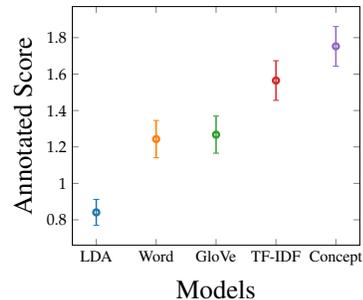

Figure 3: Annotated sentence semantic similarity scores.

It is clear that the proposed method (Concept in Fig. 3) can find more coherent sentence pairs compared to other methods. We believe this is because concept-based matching system can better distinguish concept-level similarity. For example, TF-IDF returns *So what is wrong with that argument?* as the most similar sentence to *So, what is wrong with the FDA...*, which ignores the word FDA, whereas our model returns *Look, the FDA is the biggest barrier here.* instead. LDA performs poorly because the corpus size is limited. Another interesting finding is that averaged word embeddings perform the same as the simple word overlap baseline.

## 4 CONCLUSIONS

In this work, we release a Wikipedia-based concept ontology network and describe a method to find semantically similar sentences. We further present a preliminary visualization using the proposed method to discover concept flows in debates. As for future work, we will employ entity disambiguation into this model to improve the entity detection accuracy, create an interactive visualization tool, and investigate how to model concept shifts in discourse.

## 5 ACKNOWLEDGEMENT

## REFERENCES
[1] Mauro Dragoni, Célia Da Costa Pereira, Andrea GB Tettamanzi, and Serena Villata. 2016. Smack: An argumentation framework for opinion mining. In *International Joint Conference on Artificial Intelligence (IJCAI)*. IJCAI/AAAI Press, 4242–4243.
[2] Vinodh Krishnan and Jacob Eisenstein. 2016. Nonparametric Bayesian Storyline Detection from Microtexts. In *EMNLP Workshop on Computing News Storylines*.
[3] Sana Malik, Alison Smith, Timothy Hawes, Panagis Papadatos, Jianyu Li, Cody Dunne, and Ben Shneiderman. 2013. TopicFlow: visualizing topic alignment of twitter data over time. In *ASONAM*. ACM, 720–726.
[4] Jeffrey Pennington, Richard Socher, and Christopher Manning. 2014. GloVe: Global vectors for word representation. In *EMNLP*. 1532–1543.
[5] Dafna Shahaf, Carlos Guestrin, and Eric Horvitz. 2012. Trains of thought: Generating information maps. In *WWW*. ACM, 899–908.